\documentclass[sigplan,10pt,nonacm,balance=false]{acmart}

\usepackage{amsmath}
\usepackage{booktabs}
\usepackage{algorithm}
\usepackage{algorithmic}
\newcommand{\autokernel}{\textsc{AutoKernel}}
\newcommand{\kernelarc}{\textsc{KernelArc}}
\newcommand{\tflops}{\text{TFLOPS}}

\begin{document}

\title{\kernelarc{}: A Multi-Agent Framework for GPU Kernel Optimization}

\author{Joyjit Kundu\textsuperscript{*} \quad Ben Stoffelen \quad Kaili Wang \quad Peter Vrancx \quad Ludovic Denoyer}
\affiliation{%
  \country{AILabs, Interuniversity Microelectronics Centre (IMEC)}
}

\begin{abstract}
We present \kernelarc{}, a multi-agent framework for autonomous GPU kernel optimization across heterogeneous workloads. Strategy-specialized agents run in parallel and coordinate through conclusions-only shared memory, a deterministic benchmark guard, and read-only cross-agent state with plateau-triggered drafting. We evaluate \kernelarc{} on NVIDIA H100 and B200 GPUs using category-representative SOL-ExecBench workloads. The resulting implementations span custom BF16 GEMM, static cuBLASLt Expert-API configuration tables, fused mixture-of-experts backward, shape-gated decoder-layer fusion, native NVFP4 grouped-query attention, and paged prefill attention. In the public SOL-ExecBench leaderboard snapshot recorded on August~20, 2026, \kernelarc{} ranked first on every representative L1, L2, Quantization, and FlashInfer task evaluated. The trajectories support the paper's central motivation: shared multi-agent search can broaden exploration and reach stronger incumbents within a fixed candidate budget, while the value of individual coordination features depends on the kernel and optimization stage.
\end{abstract}


\maketitle
\pagestyle{plain}
\begingroup
\renewcommand{\thefootnote}{\fnsymbol{footnote}}
\footnotetext[1]{Email: \texttt{joyjit.kundu@imec.be}}
\endgroup

\section{Introduction}
\label{sec:intro}

Extracting peak performance from modern GPU accelerators is increasingly difficult. Hopper and Blackwell expose high-throughput primitives---including Warp Group Matrix Multiply Accumulate (\texttt{wgmma.mma\_async}), Tensor Memory Accelerator (TMA) transfers, asynchronous barriers, SM100 instructions, and NVFP4 formats---but useful speedups require these mechanisms to be coordinated across layouts, the memory hierarchy, register pressure, synchronization, and launch overheads.

LLM agents offer a programmable alternative to manual performance engineering. They can profile kernels, propose structural transformations, edit CUDA/PTX or domain-specific language code, select and autotune libraries such as cuBLASLt, cuDNN, and CUTLASS, choose precision and quantization formats, and close the loop with correctness tests and benchmarks. Existing single-agent systems such as \autokernel{}~\cite{jaber2026autokernel} show that this loop can work. Their success, however, depends on a focused human-designed playbook and enough sequential budget to traverse it. Even when a single agent pivots at a plateau, it advances one incumbent under one local history, so a fixed budget covers fewer algorithm families and may over-refine a design before moving on.

We introduce \kernelarc{}, a multi-agent architecture for this fragmented-search regime. Strategy-specialized agents explore different optimization families concurrently. They exchange only validated conclusions through shared memory whose retention horizon is configurable, while a deterministic guard owns correctness, benchmarking, and keep/revert decisions. Read-only cross-agent state exposes stronger sibling solutions for inspection, and a plateau trigger asks the agent to try a different algorithm, DSL family, or data layout instead of continuing local refinements. The observed trajectories are consistent with the intended role of sharing: the configured system moves beyond a single-agent plateau, and fixed-budget ablations favor shared-memory multi-agent configurations. We report this as a system-level pattern because the value of each coordination feature can depend on the kernel and on whether search is in early exploration, plateau escape, or late local refinement. The division assigns generative reasoning to LLMs and stateful evaluation and coordination to deterministic code.

With an eight-hour wall-clock budget and a detailed Hopper-specific general matrix multiplication (GEMM) optimization playbook, a single-agent automated research loop reaches 766 $\tflops$ (BF16) on one fixed shape, 3.2\% above the matched cuBLAS baseline measured during the campaign. This demonstrates depth along one narrow playbook-guided path; it does not establish a generally faster GEMM kernel than cuBLAS or the ability to optimize a score aggregated across many shapes. Reaching 766 $\tflops$ requires a correct sequence of low-level Hopper optimizations, including WGMMA/TMA pipelining, synchronization, register-pressure management, and epilogue staging. To study the broader regime, we use the Speed-of-Light Execution Benchmark (SOL-ExecBench)~\cite{solexecbench2025}, whose SOL score aggregates performance across input shapes (higher is better). On L1-030---an attention output projection with residual addition evaluated across 16 shapes---the sequential run plateaus at SOL~0.441279, while the configured \kernelarc{} system reaches SOL~0.546380. This motivates the full shared multi-agent design without requiring every coordination feature to help at every point in the search.

We evaluate the \kernelarc{} architecture on tasks spanning all four workload categories of SOL-ExecBench~\cite{solexecbench2025}, covering common neural-network operators.
Together with the one-agent GEMM configuration, the resulting implementations span a PTX-assisted custom BF16 GEMM implementation benchmarked against a matched cuBLAS baseline, static cuBLASLt configuration tables, fused mixture-of-experts (MoE) backward, shape-gated decoder-layer fusion, native NVFP4 attention, and paged prefill attention, showing transfer across optimization axes rather than a single encoded kernel family. Detailed kernels and public-ranking outcomes appear in Appendix~\ref{sec:kernel-details}.

\noindent\textbf{Contributions.}
\begin{enumerate}
    \item \textbf{The \kernelarc{} architecture:} strategy-specialized parallel agents coordinated by conclusions-only memory with configurable retention, a deterministic benchmark guard, plateau-triggered drafting, and read-only cross-agent state; reusable conclusions remain LLM-visible, while attempt counts, stop thresholds, relaunch logic, and internal-leaderboard publication stay in deterministic code.
    \item \textbf{Evidence for shared multi-agent search:} the observed trajectories support the paper's central motivation that sharing validated conclusions can broaden search beyond isolated local trajectories, while leaving open which coordination features matter most in which kernel regimes.
    \item \textbf{Efficient kernels across distinct optimization axes:} \kernelarc{} produces implementations ranging from PTX-assisted BF16 GEMM and static cuBLASLt configuration tables to fused MoE backward, shape-gated decoder-layer fusion, native NVFP4 attention, and paged prefill attention; our SOL-ExecBench submissions ranked first on their respective tasks in the public leaderboard snapshot.\footnote{Public leaderboard ranks recorded on August~20, 2026.}
\end{enumerate}

\section{Related Work}
\label{sec:related}

\paragraph{LLM-based kernel generation.}
Several systems use LLMs for one-shot or iterative kernel generation. KernelBench~\cite{ouyang2025kernelbench} provides more than 250 GPU kernels for evaluating LLM-generated code. AutoComp~\cite{hong2025autocomp} targets portable LLM-driven kernel optimization across tensor accelerators, pairing hardware-specific optimization agents with evaluation backends for targets such as NVIDIA GPUs, TPUs, AWS Trainium, Gemmini, RISC-V vector processors, and Apple Silicon. \autokernel{}~\cite{jaber2026autokernel} introduced an autonomous agent loop: profiling a complete workload to identify bottlenecks, ranking them by Amdahl's law, and iteratively refining Triton or CUDA C++ kernels through a five-stage correctness harness and six-tier optimization playbook. In its H100 FP16 kernel benchmarks, \autokernel{} reports up to 5.29$\times$ over PyTorch eager on RMSNorm and 2.82$\times$ on softmax. \kernelarc{} studies a structurally similar observe--edit--evaluate--retain loop as its single-agent boundary case, then adds Hopper-specific guidance for the GEMM study and shared multi-agent search orchestration for broader workloads.

\paragraph{Search-based code optimization.}
Weco first introduced AIDE as a system that frames ML engineering as code optimization and formulates trial-and-error as a \emph{tree search} over candidate solutions, branching from promising nodes and pruning failures~\cite{jiang2025aide}. Weco's second report extends this line to recursive self-improvement by using an outer autoresearch loop to rewrite and evaluate the inner-loop agent harness itself~\cite{weco2026aide2}. \kernelarc{} is complementary: it keeps the agent harness, guard, and launcher fixed, and studies how shared state and strategy specialization affect GPU-kernel search under strict correctness and benchmarking constraints.

\paragraph{Multi-agent optimization architectures.}
Existing systems span flat role specialization, hierarchical archives, beam search, and generic orchestration. Astra~\cite{wei2025astra} uses five role-specialized agents without persistent memory; AKO4X~\cite{ako2026} uses cross-session archives and dead-end catalogs; AccelOpt~\cite{zhang2025accelopt} combines width-4 beam search with experience distillation; and CORAL~\cite{qu2026coral} provides worktree isolation, shared state, and heartbeat-triggered pivots. Their reported speedups are not directly comparable across tasks, hardware, or budgets. \kernelarc{} combines strategy specialization, conclusions-only memory with a configurable horizon, an external deterministic guard, and draft-on-plateau.

\paragraph{Auto-tuning and compiler optimization.}
Traditional auto-tuners and compiler stacks, such as TVM~\cite{chen2018tvm} and its Ansor auto-scheduler~\cite{zheng2020ansor}, optimize within compiler-defined schedule or IR spaces, including choices such as tiling, unrolling, vectorization, tensorization, and operator fusion. These spaces are powerful but typically do not expose arbitrary source-level rewrites, hardware-specific control-flow restructuring, or low-level CUDA/PTX idioms such as manual warp specialization. \kernelarc{} instead searches kernel source code directly.

\section{Methodology}
\label{sec:methodology}

\subsection{Measurement-Gated Kernel Optimization}

Autonomous kernel optimization starts from a seed implementation $k_0$, a workload set $W$, a target hardware/software environment $H$, and a deterministic evaluator. The search objective is
\begin{equation}
    k^\star = \arg\max_{k \in \mathcal{K}} S(k; W,H)
    \quad \text{s.t.} \quad C(k,w)=1 \;\; \forall w \in W,
\end{equation}
where $\mathcal{K}$ is the space of source-level implementations, $C$ is the correctness predicate, and $S$ is a benchmark score, such as the inverse of the geometric mean latency across all workloads of a given problem. The constraint $C(k,w){=}1$ requires candidate $k$ to pass the evaluator's correctness check on workload case $w$; candidates that fail any $w \in W$ are infeasible regardless of speed. Here, $\mathcal{K}$ includes structural changes---for example, replacing a framework operator with a vendor-library call, changing the kernel DSL, fusing an epilogue, or introducing warp specialization---in addition to parameters from a fixed schedule template.

An agent observes the current implementation and measurements, forms a bottleneck hypothesis, edits the kernel, and submits the candidate to the evaluator. Correct candidates that improve the incumbent are retained as the accepted best; non-improving candidates do not ratchet that best, although \kernelarc{} can temporarily keep near-best lateral candidates as working points. The evaluator must run in the target environment because compiler versions, library heuristics, clocks, cache state, and workload aggregation all affect candidate ordering.

\subsection{Guarded Single-Agent Optimization Loop}

The single-agent boundary case of \kernelarc{} follows a measurement-gated loop, structurally similar to prior autonomous kernel-optimization systems such as \autokernel{}~\cite{jaber2026autokernel}: observe the incumbent, form a bottleneck hypothesis, edit one kernel, run correctness and timing checks, then ratchet the accepted best only on measured improvements. This loop is useful for studying depth: it refines one incumbent under one local history and exposes how far a focused trajectory can go. Its conclusions are therefore tied to the workload, shape, call pattern, model backbone, and budget. Broader coverage across shapes, operators, and search directions motivates \kernelarc{}'s shared multi-agent design.

\subsection{From Sequential Plateaus to Design Requirements}
\label{sec:wall}

Sequential search may become a search-process bottleneck when the target is no longer one fixed-shape operator but a multi-shape SOL-ExecBench task. For L1-030, an attention output projection with residual addition on Blackwell B200, the submission is scored over \textbf{16 input shapes}; a good search must therefore combine per-shape configuration selection, fusion, precision choices, and faithful cold-cache measurement.

\paragraph{Why one trajectory plateaus.}
Different L1-030 shapes can favor different library configurations, precision modes, and fusion strategies. A sequential agent can investigate only one active direction at a time and may spend part of its budget revisiting nearby configurations after saturation. Its local history can therefore mistake a search-trajectory ceiling for a hardware ceiling, especially when other implementation families remain unexplored.

This failure yields four system requirements: diversify optimization directions, transfer validated lessons without transferring entire histories, keep evaluation and coordination state deterministic, and provide explicit escape routes from local plateaus.

\subsection{\kernelarc{} Architecture}
\label{sec:kernelarc}

\begin{figure}[t]
\centering
\includegraphics[width=\linewidth]{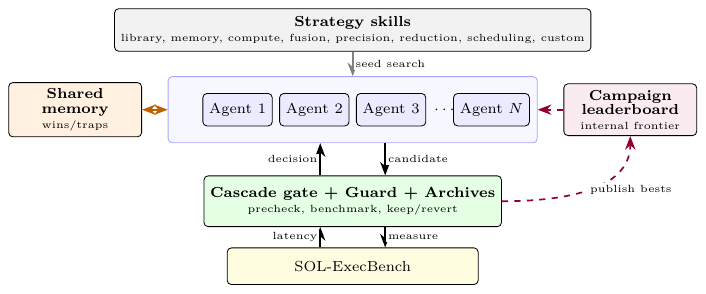}
\caption{\kernelarc{} architecture. Strategy skills seed the agent portfolio; agents recall and record validated conclusions through shared memory, then submit candidates to a deterministic cascade gate and guard. The guard benchmarks candidates on SOL-ExecBench, returns REJECT/KEEP/ACCEPT/REVERT actions and STOP feedback, archives kept best variants, and exposes guard bests through an internal campaign leaderboard for read-only cross-agent inspection. This campaign leaderboard is distinct from the public NVIDIA SOL-ExecBench leaderboard used for external rankings.}
\Description{Strategy skills seed a portfolio of agents. Agents exchange wins and traps through shared memory, submit candidates to a cascade gate and guard, receive REJECT, KEEP, ACCEPT, REVERT, or STOP feedback, and use SOL-ExecBench measurements for validation. The guard may publish an internal campaign leaderboard for read-only cross-agent inspection.}
\label{fig:kernelarc}
\end{figure}

\kernelarc{} maps the requirements to configurable mechanisms: specialized agents diversify the search; memory distills transferable conclusions with an adjustable retention horizon; the cascade gate and guard own validation, evaluation, and keep/revert state; and plateau drafting can optionally use read-only cross-agent state derived from guard archives. Figure~\ref{fig:kernelarc} shows the evaluated architecture. Section~\ref{sec:ablations} evaluates complete configurations rather than marginal contributions of individual mechanisms.

\paragraph{Strategy-specialized parallel search.}
\kernelarc{} exposes eight strategy skills: overlapping search lenses that emphasize different parts of the optimization space rather than mutually exclusive directions. Each skill is a lightweight seed document with YAML frontmatter (name, description, eligible kernel types) and representative optimization patterns. Different skills are intended to keep concurrent agents from collapsing onto the first plausible family. Skills seed directions rather than prescribe solutions: agents derive implementations, parameters, and cross-strategy combinations. A meta-skill (\emph{playbook-index}) records eligibility constraints. Table~\ref{tab:skills} summarizes the skill set used to seed search; meta-skills specify how strategies are selected, remembered, and benchmarked.

\begin{table}[t]
\centering
\small
\caption{KernelArc skills and meta-skills.}
\label{tab:skills}
\begin{tabular}{ll}
\toprule
\textbf{Skill} & \textbf{Role} \\
\midrule
\texttt{strat-library} & cuBLASLt\slash cuDNN paths \\
\texttt{strat-memory} & coalescing, cache reuse \\
\texttt{strat-compute} & tensor-core use \\
\texttt{strat-fusion} & epilogue fusion \\
\texttt{strat-precision} & FP8\slash BF16 choices \\
\texttt{strat-reduction} & softmax, normalization \\
\texttt{strat-scheduling} & persistent kernels, split-K \\
\texttt{strat-custom-b200} & SM100-native kernels \\
\midrule
\texttt{playbook-index} & strategy eligibility \\
\texttt{memory} & shared wins and traps \\
\texttt{sol-benchmark} & benchmark protocol \\
\bottomrule
\end{tabular}
\end{table}

An eligibility map prevents agents from applying irrelevant strategies; for example, library and memory strategies apply broadly, while SM100-native custom kernels are enabled only for GEMM-like tasks.

\paragraph{Conclusions-only memory as a retention knob.}
When shared memory is enabled, agents read from and write to a common memory store under advisory locking to avoid write races. The memory contains:

\begin{itemize}
    \item \textbf{Wins}: Successful optimizations with speedup, shape context, and a \emph{reflect} field (``WHY it worked + before$\to$after pseudocode''). The default configuration retains 16 per kernel type, sorted by speedup descending.
    \item \textbf{Traps}: Dead ends with an error description and reflect field (``WHY it failed''). The default configuration retains 16 per kernel type in first-in, first-out (FIFO) order.
\end{itemize}

The store excludes iteration counters, heartbeats, and progress logs. Thus shared memory carries reusable conclusions, while counters, accepted archives, stop conditions, restarts, and campaign-leaderboard publication remain in deterministic code. Operational traces show accumulated traps progressively narrowing the active strategy set. For long runs, a finite retention horizon may keep recall compact and reduce context pollution from stale or low-value entries; the horizon remains a configuration knob, and the present experiments do not prove that bounded retention is better than full history. Field limits (e.g., 160 characters for strategy and 1200 for reflection) prevent verbose entries from dominating recall.

\paragraph{External deterministic guard.}
A deterministic process 
mediates all kernel changes, so the agent proposes code but never decides whether its own edit is an improvement:

\begin{enumerate}
    \item \textbf{Cascade gate}: Cheap pre-benchmark validation (syntax, \texttt{run()} entry point, portability---no \texttt{ctypes}\slash\texttt{subprocess}\slash\texttt{cpp\_extension} for Python kernels).
    \item \textbf{Benchmark}: Execute via the SOL-ExecBench bridge under a GPU lock; measure the configured aggregate latency across workloads.
    \item \textbf{Keep condition}: mark a candidate as KEEP only if it passes every workload and improves the incumbent by the configured fractional margin.
    \item \textbf{Stop conditions}: guard STOP is emitted on plateau, target reached, time budget exhausted, or a configured candidate cap.
\end{enumerate}

On KEEP, the kernel is snapshotted to the guard's best and accepted archives. Near-best lateral candidates may be left in the working directory for exploration without ratcheting the best score, but still count as non-improvements for plateau tracking; on REVERT, the previous best is restored.
The cascade gate rejects malformed or prohibited Python submissions before invoking the GPU benchmark, avoiding unnecessary SOL-ExecBench runs; we use this as a validation safeguard rather than treating it as an isolated mechanism estimate.

\paragraph{Iteration protocol and plateau escape.}
\label{sec:draft-on-plateau}
Each agent follows the loop in Algorithm~\ref{alg:kernelarc-agent}, while Algorithm~\ref{alg:kernelarc-launcher} starts the agent portfolio and maintains shared state. In each iteration, an agent recalls relevant wins and traps, optionally inspects faster sibling archives read-only, chooses an eligible strategy, edits its own solution directory, and submits the candidate to the guard. The guard returns one of four outcomes: REJECT for cheap pre-benchmark failures, KEEP for a measured improvement, ACCEPT for a correct near-best candidate that remains as the working point while the best score is unchanged, and REVERT for a regression or benched failure. The plateau counter $r$ resets on KEEP, advances only on benched non-improvements (ACCEPT or REVERT), and ignores cheap REJECTs. Once $r$ reaches the drafting threshold $r_\mathrm{draft}$, the agent stops making local tweaks and instead requests a different algorithm, DSL family, or data layout.

\begin{algorithm}[t]
\caption{\kernelarc{} Agent Optimization Loop}
\label{alg:kernelarc-agent}
\begin{algorithmic}[1]
\REQUIRE Problem $P$, solution dir $d$, starting DSL, strategy set $S$, memory $M$, guard $G$
\STATE Bootstrap $d$ from $P$; initialize $G$ with a baseline benchmark and archive
\STATE Determine kernel type and eligible strategies $S$
\STATE $r \leftarrow 0$ \COMMENT{consecutive benched non-improvements}
\REPEAT
    \STATE \textbf{Recall}: read wins/traps for the kernel type and inspect own archive of kept bests
    \STATE \hspace{1em} If enabled, read internal campaign leaderboard $L$ and inspect only listed sibling archives (read-only)
    \IF{$r \geq r_\mathrm{draft}$ and drafting is enabled}
        \STATE \textbf{Draft}: request a fundamentally different DSL, algorithm, or data layout
    \ELSE
        \STATE \textbf{Plan}: choose an eligible strategy $s \in S$ and form a concrete hypothesis
    \ENDIF
    \STATE Generate candidate edits in $d$ that write into benchmark-provided output tensors
    \FOR{each candidate}
        \STATE \textbf{Verify}: $(\mathit{action}, \tilde{m}) \leftarrow G.\text{check}(d)$ \COMMENT{REJECT, KEEP, ACCEPT, or REVERT}
        \IF{$\mathit{action} = \text{KEEP}$}
            \STATE $r \leftarrow 0$; $M.\text{record\_win}(\text{type}, s, \text{speedup}, \text{reflect})$
        \ELSIF{$\mathit{action} = \text{ACCEPT}$}
            \STATE $r \leftarrow r + 1$; keep the lateral candidate as working code; best archive unchanged
        \ELSIF{$\mathit{action} = \text{REVERT}$}
            \STATE $r \leftarrow r + 1$; restore the best archive; record trap or repair/pivot when fundamental
        \ELSE
            \STATE Record cheap rejection/trap as appropriate; $r$ unchanged \COMMENT{REJECT}
        \ENDIF
    \ENDFOR
\UNTIL{$G$ emits \texttt{STOP} or the configured round/strategy budget is exhausted}
\STATE If producing a submission, pack the best accepted kernel via $G.\text{pack}(d)$
\end{algorithmic}
\end{algorithm}

\begin{algorithm}[t]
\caption{\kernelarc{} Launcher Loop}
\label{alg:kernelarc-launcher}
\begin{algorithmic}[1]
\REQUIRE Problem $P$, solution-dir base $d_0$, starting DSL, strategies $S$, memory mode, campaign-leaderboard flag
\STATE Split $S$ into one or more agent strategy groups (or assign all strategies to one agent)
\STATE Configure shared or private memory paths; allocate one solution dir $d_i$ per agent
\FOR{each agent group $i$}
    \STATE Spawn agent $a_i$ with $P$, $d_i$, starting DSL, assigned strategies, memory path, and guard-only GPU access
\ENDFOR
\IF{campaign leaderboard enabled}
    \STATE Start one publisher that scans exactly $\{d_i\}$ and atomically writes $L$ every $\Delta$ seconds
\ENDIF
\WHILE{keep-alive enabled and some guard status has not emitted \texttt{STOP}}
    \FOR{each agent $a_i$}
        \IF{$a_i$ exits before $G_i$ reports \texttt{STOP}}
            \STATE Reinvoke $a_i$ on the same $d_i$; the guard archive preserves the best accepted variant
        \ENDIF
    \ENDFOR
\ENDWHILE
\STATE Per-agent bests remain in guard archives; the campaign leaderboard and ledgers expose the internal frontier
\end{algorithmic}
\end{algorithm}

\section{Experimental Results}
\label{sec:experiments}

\subsection{Guided Sequential GEMM Boundary Study}
\label{sec:sequential}

\paragraph{Role in the paper.}
This study is a deliberately favorable boundary case for the one-agent, private-memory limit of \kernelarc{}, not a claim that generated kernels generally outperform vendor libraries. It asks how far one agent can go when a human-designed playbook already identifies the relevant optimization path. The result establishes achievable \emph{depth}; its shape specificity, call-pattern dependence, guidance burden, and sequential plateau illustrate the depth regime behind \kernelarc{}'s shared multi-agent emphasis on search \emph{breadth}.

\paragraph{GEMM campaign setup.}
We instantiate this setup with a 4096$\times$4096 BF16 matrix multiplication on an NVIDIA H100 SXM (132 streaming multiprocessors (SMs), HBM3 at 3.35\,TB/s, and approximately 989 $\tflops$ BF16 Tensor Core peak). The agent begins from a textbook-na\"ive CUDA kernel and receives an eight-hour wall-clock budget. During that campaign, we compare each accepted candidate with the strongest cuBLAS baseline measured under the same end-to-end protocol using \texttt{cublasGemmEx}. Because cuBLAS throughput depends on library version, selected algorithm, preprocessing policy, and measurement protocol, the comparison is specific to this repeated-call environment.

\paragraph{Playbook and kernel provenance.}
The one-agent \kernelarc{} configuration is the framework's private-memory boundary case, using a measurement-gated loop of the same general form as prior autonomous kernel optimizers such as \autokernel{}~\cite{jaber2026autokernel}. Following \autokernel{}'s playbook formulation, this campaign specializes the guidance to Hopper-specific mechanisms centered on WGMMA/TMA, warp specialization, explicit barriers, epilogue design, and cache policy. Table~\ref{tab:tiers} summarizes the CUDA curriculum used in the campaign. The playbook is detailed rather than a short hint list: it specifies a dependency-ordered tier sequence, architectural prerequisites, profiler targets, and failure modes, but not the final kernel or exact schedule. The agent remains responsible for translating the curriculum into correct CUDA/PTX, selecting tile and pipeline parameters, debugging compiler and synchronization failures, composing techniques, and exploring campaign-specific refinements not prescribed verbatim---including descriptor caching, wait-depth relaxation, a chunked TMA epilogue, parallel issuance of A/B TMA loads, and zero-stack K-loop unrolling. No human edits are applied to candidate kernel code during the reported run; the 766 $\tflops$ kernel is the direct output of the agent loop.

\begin{table}[t]
\centering
\caption{Condensed Hopper campaign curriculum. Rows follow the CUDA playbook dependency chain and summarize the campaign qualitatively, not as isolated ablations.}
\label{tab:tiers}
\scriptsize
\begin{tabular}{@{}p{0.07\columnwidth}@{\hspace{0.5em}}p{0.53\columnwidth}@{\hspace{0.5em}}p{0.32\columnwidth}@{}}
\toprule
\textbf{Tier} & \textbf{Curriculum Stage} & \textbf{Role in Campaign} \\
\midrule
1 & Scalar warp-tiled CUDA baseline & Correct tiled starting point \\
2 & WGMMA/TMA tensor-core pipeline & Main compute transition \\
3 & Producer/consumer warp specialization & Overlap loads and compute \\
4 & Persistent CTAs and raw \texttt{mbarrier} & Control residency and phases \\
5 & Cluster-aware TMA multicast path & Shared-data reuse path \\
6 & Epilogue and wait-depth refinements & Clean up completion path \\
7 & Tile scheduling, L2 policy, and TMA issue order & Locality and final scheduling \\
\bottomrule
\end{tabular}
\end{table}

\begin{figure}[t]
\centering
\includegraphics[width=\linewidth]{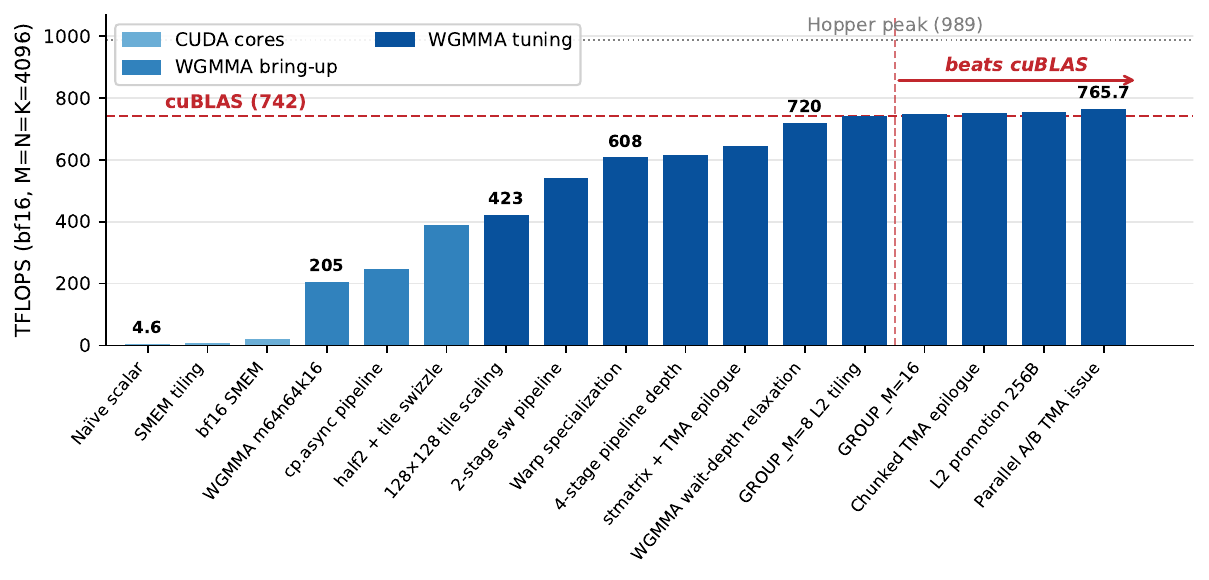}
\caption{Motivating eight-hour single-agent progression: 4.6 to 766 $\tflops$ across 17 accepted techniques on Hopper. The dashed line marks the strongest cuBLAS end-to-end baseline measured under the same protocol during the campaign (742 $\tflops$).}
\Description{A line chart rises from 4.6 to 766 TFLOPS over 17 accepted kernel variants and crosses the 742-TFLOPS cuBLAS baseline near the end.}
\label{fig:cublas_journey}
\end{figure}

\paragraph{Accepted trajectory.}
The accepted path is cumulative rather than a single final trick. Moving from scalar/data-reuse kernels to the tensor-core substrate raises throughput from 19.3 to 541 $\tflops$ across WGMMA, asynchronous-copy, tile-scaling, and software-pipeline milestones. Producer--consumer warp specialization reaches 617 $\tflops$ by dedicating one warp group to TMA loads and two to WGMMA consumers, coordinated through asynchronous \texttt{mbarrier} phase tracking. Epilogue and wait-depth changes reach 720 $\tflops$, and the final 46 $\tflops$ comes from campaign-specific refinements: descriptor and transposed-matrix caching under the repeated-call harness, zero-stack K-loop unrolling, relaxed WGMMA wait depth, and L2 promotion with 256-byte sector hints. The resulting 766 $\tflops$ is 3.2\% above the strongest cuBLAS baseline measured under the same fixed-shape protocol during the eight-hour campaign (742 $\tflops$) and reaches 77\% of theoretical peak. These intervals describe one accepted path, not component ablations, and the data do not establish a faster standalone GEMM mainloop. Figure~\ref{fig:cublas_journey} summarizes the 17 accepted milestones.

\begin{figure}[t]
\centering
\includegraphics[width=\linewidth]{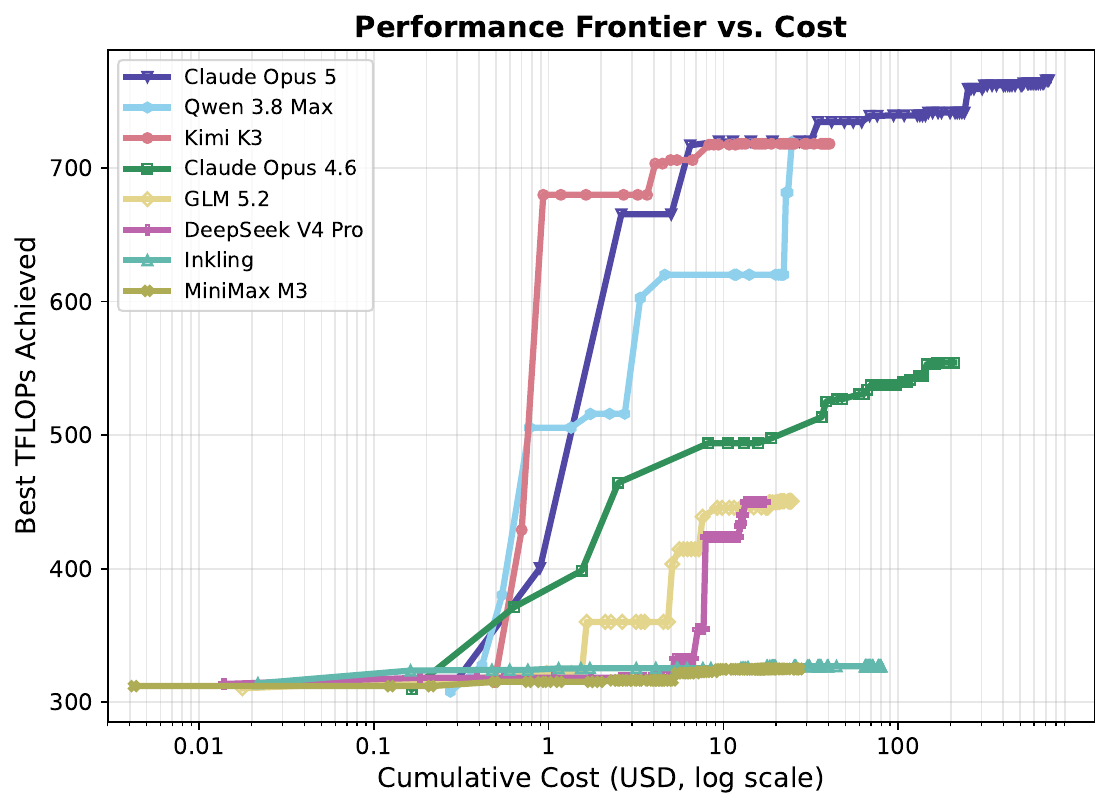}
\caption{Cost--performance frontier for eight model backbones in the motivating GEMM study (best-so-far throughput versus cumulative cost; logarithmic cost axis). Claude Opus 5 attains the highest throughput (${\sim}766\,\tflops$) but at the largest cumulative cost, while Kimi K3 reaches the near-cuBLAS regime most cheaply; the remaining backbones saturate at lower throughput.}
\Description{Eight best-so-far throughput curves versus cumulative dollar cost; Claude Opus 5 reaches the highest throughput, while Kimi K3 approaches cuBLAS at the lowest cost.}
\label{fig:cost_performance}
\end{figure}

\paragraph{Backbone cost efficiency.}
Figure~\ref{fig:cost_performance} adds a cost view from one trajectory per backbone. Claude Opus 5 reaches the campaign-best ${\sim}766\,\tflops$ after substantial spend; within this run budget, Kimi K3 is the most cost-efficient observed trajectory, exceeding 700 $\tflops$ after a few dollars before plateauing near 718 $\tflops$. Other backbones saturate lower. Thus the playbook defines a productive but narrow direction, while model choice controls its cost and depth.

\subsection{Evaluation on SOL-ExecBench}
\label{sec:results}
SOL-ExecBench~\cite{solexecbench2025} organizes its 235 problems into four categories by complexity and precision~\cite{solexecbench2025}: \emph{L1} single-operation building blocks such as GQA, RMSNorm, SwiGLU, and RoPE; \emph{L2} complete fused blocks, reported as 3--10$\times$ more complex than L1; \emph{Quantization (Q)} explicit FP8 blockwise and NVFP4 block-scaled computation; and \emph{FlashInfer-Bench (FI)} standalone inference primitives such as fused attention, FP8 MoE, and RMSNorm. Each task is evaluated over multiple input shapes and summarized by a single SOL score (higher is better, with 1 representing the hardware limit), with SM clocks locked, L2 flushed, and CUDA Profiling Tools Interface (CUPTI) tracing on Blackwell B200. We evaluate one task from each category, not the full 235-task suite, primarily because of access and cost constraints. 

\paragraph{Case study: L1-030 optimization trajectory.}

\begin{figure}[t]
\centering
\includegraphics[width=\linewidth]{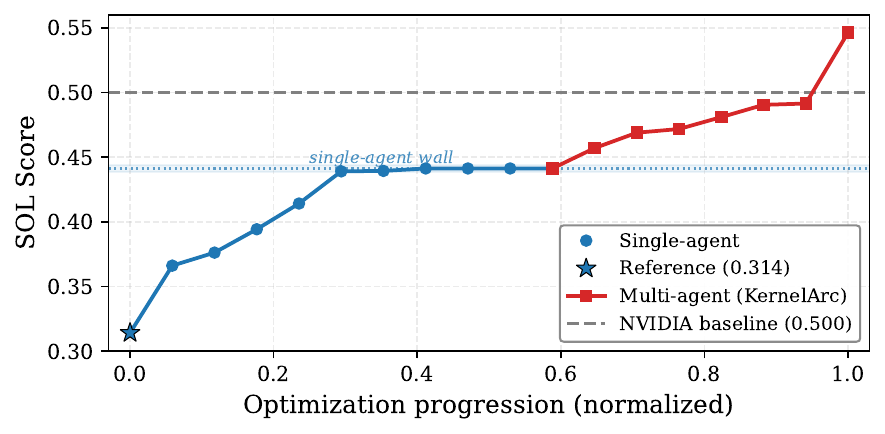}
\caption{L1-030 SOL progression using Claude Opus~4.6. Single-agent search plateaus at 0.441279; multi-agent search reaches 0.546380, exceeding NVIDIA's 0.500 baseline.}
\Description{A progression chart shows the single-agent score plateauing at 0.441279 and the multi-agent score increasing through successive submissions to 0.546380 and exceeding the NVIDIA baseline.}
\label{fig:sol_progression}
\end{figure}
The single-agent search reached SOL~0.441279 and plateaued after consecutive reverts; the multi-agent search escaped this local plateau and advanced from 0.457047 through intermediate bests of 0.490500 and 0.491515 to 0.546380, eventually exceeding the 0.500 NVIDIA baseline.
The performance improvement of L1-030 happens gradually through library-configuration, fusion, and measurement refinements rather than one large algorithmic substitution.

The final kernel reflects this pattern: it aliases the output tensor onto the residual buffer so cuBLASLt folds residual addition into the GEMM epilogue, then uses a static per-$M$ table of validated cuBLASLt Expert-API settings with heuristic fallback. The artifact is library configuration plus residual aliasing, not a handwritten GEMM mainloop.

\paragraph{Cross-category evaluation across L1, L2, Q, and FI.}
\label{sec:cross_benchmark}
We deployed \kernelarc{} on L1-030 (GEMM+residual; Claude Opus~4.6), L2-025 (MoE backward; Kimi K3), L2-053 (decoder layer; Kimi K3), Quant-031 (NVFP4 grouped-query attention; Kimi K3), and FI-014 (paged attention; Kimi K3). Including the square-GEMM kernel discussed above, these studies span PTX-assisted custom kernel implementation, library selection, fusion, bandwidth optimization, quantized tensor-core throughput, and latency-sensitive attention; they are case studies, not a suite-wide win-rate estimate. Table~\ref{tab:leaderboard-latency-speedups} reports the corresponding public-leaderboard latency-ratio speedups for four representative problems. The guard, memory interface, launcher, and agent protocol remain fixed; only taxonomy-driven strategy eligibility changes. Appendix~\ref{sec:kernel-details} gives the cross-category comparison, public-ranking snapshot, and kernel details.

\begin{table}[t]
\centering
\small
\caption{Latency-ratio speedups for this work in the August~20, 2026 public NVIDIA SOL-ExecBench leaderboard snapshot\protect\footnotemark{}. Values are computed as baseline latency divided by this work's latency.}
\label{tab:leaderboard-latency-speedups}
\begin{tabular}{lcc}
\toprule
Problem & vs. PyTorch reference & vs. optimized baseline \\
\midrule
L1-030 & 1.40$\times$ & 1.06$\times$ \\
L2-025 & 291.44$\times$ & 1.13$\times$ \\
Quant-031 & 1327.04$\times$ & 43.79$\times$ \\
FI-014 & 61972.57$\times$ & 143.78$\times$ \\
\bottomrule
\end{tabular}
\end{table}
\footnotetext{\url{https://research.nvidia.com/benchmarks/sol-execbench/leaderboard}}

\subsection{Ablation Studies}
\label{sec:ablations}

\paragraph{Scope and configurations.}
Repeated end-to-end ablations are expensive because every trajectory consumes both B200 time and model inference. We therefore study one representative workload, FI-014, the causal paged-GQA prefill task (\href{https://research.nvidia.com/benchmarks/sol-execbench/kernel/223}{SOL-ExecBench kernel 223}), using Kimi K3 throughout; among our representative tasks, FI-014 showed the largest observed headroom relative to both the reference implementation and scoring baseline. We compare three directory-defined configurations with five trajectories each: \emph{Single} uses private memory and no cross-agent campaign leaderboard; \emph{Multi-bounded} launches two agents with shared win/trap memory capped at 16 entries and read-only campaign-leaderboard access; and \emph{Multi-unbounded} uses the same two-agent shared configuration without a memory cap. Each trajectory has a total budget of 100 candidate positions, split evenly across agents in the multi-agent configurations. We also ran pilot two- and four-agent variants. Under the fixed 100-candidate budget, increasing the portfolio width from two to four agents did not show a consistent additional gain on FI-014; we therefore report the two-agent shared-memory arms as the controlled comparison and treat agent-count scaling as a separate question. The task, B200 target, strategy roster, guard, and keep/revert rule are otherwise fixed.

\paragraph{Fixed-budget analysis.}
Every attempted candidate consumes one position, including crashes and rejected candidates; only correct candidates retained by the guard can improve the incumbent. Five trajectories stop before position 100, so their last observed incumbent is carried forward by at most 23 positions. This avoids crediting unobserved improvements while comparing all runs at a common budget. Because speedups are positive ratios normalized to each trajectory's starting seed, we report geometric means with trajectory-bootstrap confidence intervals.

\begin{figure}[t]
\centering
\includegraphics[width=\linewidth]{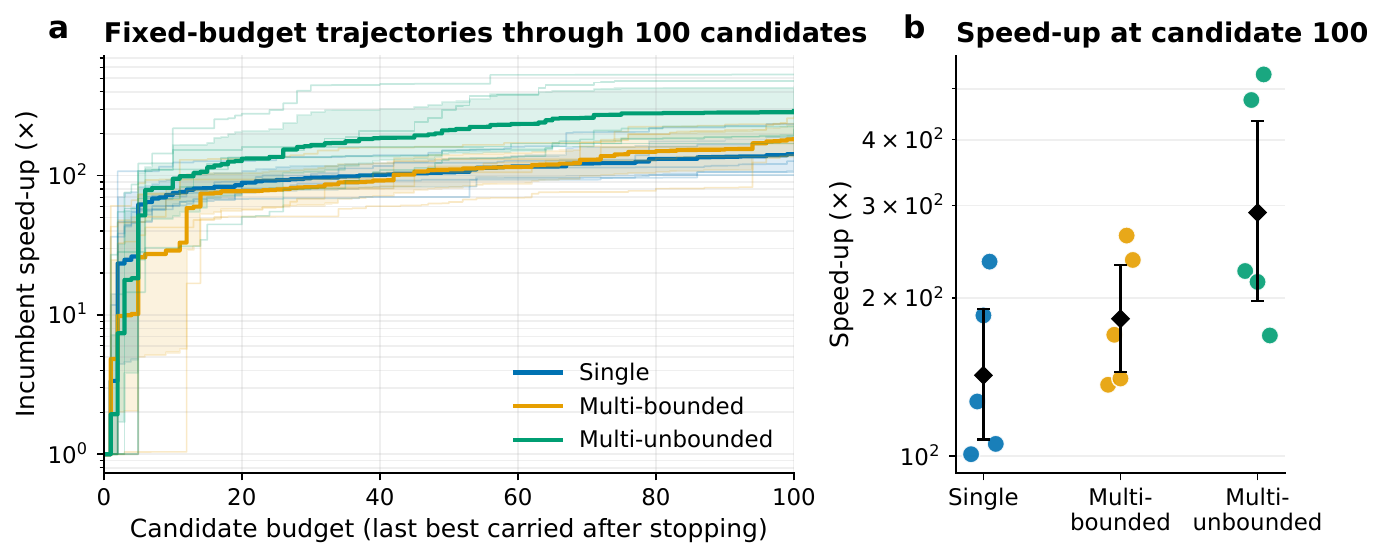}
\caption{Fixed-budget optimization on FI-014 over five trajectories per configuration. (a)~Faint step lines are individual incumbent-speedup trajectories; thick lines are category geometric means, and shaded regions are pointwise 95\% trajectory-bootstrap intervals. (b)~Candidate-100 outcomes, with individual trajectories overlaid on the geometric mean and 95\% bootstrap interval (both panels use a log y-axis).}
\Description{Two panels compare single, multi-agent bounded-memory, and multi-agent unbounded-memory optimization. Individual trajectories and geometric-mean confidence bands are shown through 100 candidate positions; endpoint distributions show the unbounded-memory configuration with the highest geometric mean and greatest dispersion.}
\label{fig:ablation}
\end{figure}

\paragraph{Results.}
Table~\ref{tab:fi014-ablation-results} summarizes the endpoint in Figure~\ref{fig:ablation} over five trajectories per configuration. Speedups are normalized to each trajectory's starting seed; across all ablation runs, the starting seeds have an average latency of 2.097~ms (reference PyTorch implementation: 345.868913~ms). At a roughly 100-candidate budget, Multi-unbounded is the strongest observed configuration: its geometric-mean speedup is 2.04$\times$ Single and 1.59$\times$ Multi-bounded, while Multi-bounded is 1.28$\times$ Single. The trajectory view shows the same ordering, although configurations overlap early and individual runs plateau at different levels.

\begin{table}[t]
\centering
\small
\caption{FI-014 ablation at a roughly 100-candidate budget. Speedups and latencies are geometric means with 95\% trajectory-bootstrap intervals.}
\label{tab:fi014-ablation-results}
\begin{tabular}{lcc}
\toprule
Configuration & Speedup ($\times$) & Latency (ms) \\
\midrule
Single & 142.6 [107.6, 190.8] & 0.0132 [0.0116, 0.0153] \\
Multi-bounded & 182.7 [144.5, 231.0] & 0.0092 [0.0076, 0.0113] \\
Multi-unbounded & 290.8 [197.3, 434.2] & 0.0085 [0.0057, 0.0130] \\
\bottomrule
\end{tabular}
\end{table}

\paragraph{Uncertainty and interpretation.}
The updated five-seed analysis strengthens the Multi-unbounded contrast, but inference remains exploratory. Its geometric standard-deviation factor is 1.67, compared with 1.35 for Multi-bounded and 1.44 for Single. This dispersion is expected in source-code search: early kept edits can steer later proposals toward different local regions of the implementation space, so a few trajectories may end with substantially different incumbents under the same budget. The exact permutation test on log speedup gives $p{=}0.0466$ in the primary grouping, but excluding the legacy non-conforming Single trajectory weakens it to $p{=}0.0906$ while preserving the qualitative ordering. The configurations also change several factors together---agent count, memory scope, campaign-leaderboard access, and retention horizon---so these data support a workload-specific trend toward stronger FI-014 incumbents under shared multi-agent search, not a universal effect size or a causal estimate for any single knob.

\section{Analysis and Discussion}
\label{sec:analysis}

\paragraph{Search phases and conclusions-only memory.}
Across campaigns, early exploration uses eligible skills as diverse optimization lenses; accepted changes then establish a few viable implementation families; and mature runs often revisit similar schedules---search-process fixed points---where improvements become local configuration, launch, or data-movement refinements. The FI-014 ablation follows this pattern: isolated extra trajectories add little over Single on average, whereas shared-memory multi-agent runs reach stronger incumbents at the same candidate budget. Shared traps can prevent repeated dead ends, and shared wins can redirect nearby search. \kernelarc{} makes these conclusions reusable through field limits, configurable retention, and canonical validation. Compact retention may reduce context pollution in long runs; full history may preserve rare useful observations. The present experiments treat this retention horizon as a knob and do not establish which setting is generally preferable. Prior systems also persist experience~\cite{ako2026,zhang2025accelopt,qu2026coral}, but cross-system comparisons confound memory with topology, model, budget, and benchmark.

\paragraph{Playbook depth versus breadth.}
The one-agent Hopper configuration is deep and prescriptive for one GPU family; \kernelarc{}'s shared multi-agent strategies are broad, lightweight optimization axes. The results illustrate the corresponding regimes: detailed guidance takes one fixed GEMM to 766 $\tflops$, whereas multi-shape tasks require coverage across operator- and shape-dependent directions. Both playbooks guide search without supplying final implementations.

\paragraph{Limitations.}
The case studies do not estimate a suite-wide win rate across all 235 problems, and rankings are time-specific. The Hopper--Blackwell contrast changes both workload and hardware. The ablation has few heterogeneous runs and uses evaluated candidates rather than wall time, tokens, or dollar cost. Isolating each coordination feature is difficult because the feature that helps may depend on both the kernel and the optimization regime---early diversification, plateau escape, or late-stage refinement.

\section{Conclusion}
\label{sec:conclusion}

\kernelarc{} studies GPU-kernel optimization as coordinated, measurement-gated search. Its strategy skills, deterministic guard, and reusable win/trap memory let agents explore different optimization lenses while sharing validated conclusions. Across the evaluated SOL-ExecBench tasks, the same framework supports library-backed kernels, fused operators, quantized attention, paged prefill attention, and custom GEMM development. The ablation study supports the coordination hypothesis: at a roughly 100-candidate budget, shared-memory multi-agent runs outperform Single on average, with Multi-unbounded reaching 2.04$\times$ the Single geometric-mean speedup. The one-agent GEMM study shows the complementary depth regime, reaching 766 $\tflops$ on one repeated-call shape when a clear hardware-specific playbook is available. Overall, the results suggest that measured coordination can help kernel search, while per-mechanism causal attribution and suite-wide effect sizes remain open for larger controlled studies.

\appendix
\section{Cross-Category Kernel and Public-Ranking Details}
\label{sec:kernel-details}

\begin{figure}[t]
\centering
\includegraphics[width=\linewidth]{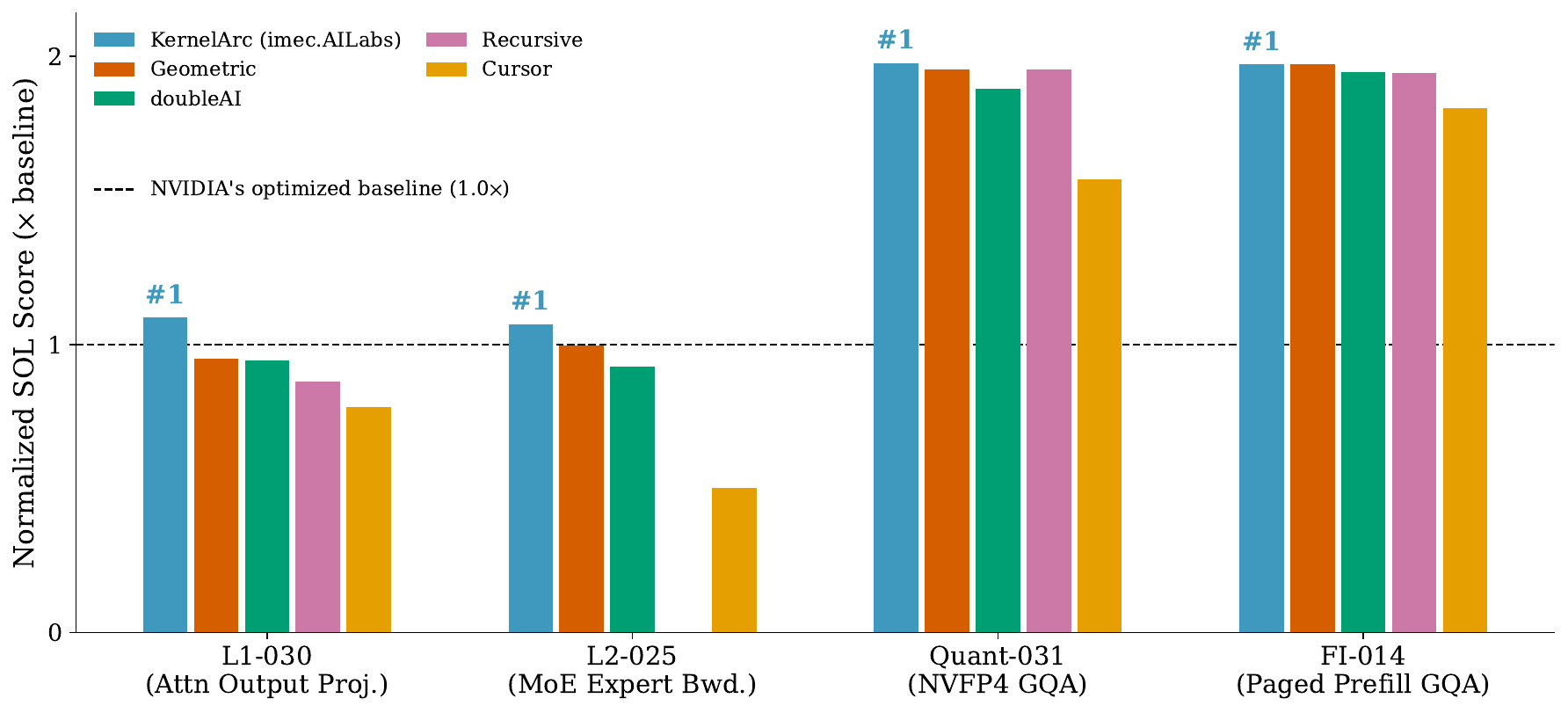}
\caption{Baseline-normalized public SOL scores for the selected L1, L2, Q, and FI tasks (task SOL divided by NVIDIA baseline SOL). Values above 1.0 outperform the optimized baseline. Each of the four \kernelarc{} submissions shown ranked first in the public leaderboard snapshot.\protect\footnotemark{}}
\Description{A grouped bar chart shows KernelArc above the NVIDIA baseline on selected L1, L2, Quantization, and FlashInfer tasks.}
\label{fig:sol_cross_benchmark}
\end{figure}
\footnotetext{Public leaderboard snapshot recorded on August~20, 2026.}
\newcounter{leaderboardranknote}
\setcounter{leaderboardranknote}{\value{footnote}}

\paragraph{Public-ranking snapshot.}
In the public leaderboard snapshot, L1-030, L2-025, L2-053, Quant-031, and FI-014 ranked first\footnotemark[\value{leaderboardranknote}] on their respective tasks. Rankings are time-specific, and these tasks do not estimate a suite-wide win rate.

\paragraph{L1-030: attention output projection with residual.}
The archived L1-030 solution is a CUDA extension that aliases the destination-passing output onto the residual tensor, making the cuBLASLt epilogue compute $D=\alpha W^\top A+\beta C$ with $C{=}D$ and $\beta{=}1$. The implementation does not autotune during timed evaluation; it constructs a cached per-$M$ cuBLASLt plan from a validated Expert-API table, with one measured heuristic-rank override and a heuristic fallback if no checked table entry is usable.

\begin{algorithm}[t]
\caption{L1-030 static cuBLASLt residual epilogue}
\label{alg:l1-030-pseudocode}
\begin{algorithmic}[1]
\REQUIRE attention output $A$, residual $R$, projection weight $W$, destination tensor $O$
\STATE $M \leftarrow$ flattened token count of $A$
\STATE $P \leftarrow$ lookup cached cuBLASLt plan for $M$
\IF{$P$ is not cached}
    \STATE $E \leftarrow$ lookup static Expert-API table entry for $M$
    \IF{$M{=}512$ and the validated heuristic-rank-1 plan is available}
        \STATE $P \leftarrow$ that heuristic cuBLASLt plan
    \ELSIF{$E$ exists and \textsc{AlgoCheck} accepts the cuBLASLt algorithm-66 configuration specified by $E$}
        \STATE $P \leftarrow$ Expert-API plan from $E$
    \ELSE
        \STATE $P \leftarrow$ heuristic-rank-0 cuBLASLt fallback plan
    \ENDIF
\ENDIF
\STATE alias $O$ to the storage of $R$ so that $C$ and $D$ are the same buffer
\STATE run cuBLASLt matmul $O \leftarrow W^\top A + O$ with $\alpha{=}1,\beta{=}1$
\RETURN
\end{algorithmic}
\end{algorithm}
\noindent Final SOL: \textbf{0.546380} (1st\footnotemark[\value{leaderboardranknote}], 14/16 shapes faster than baseline, 1.06$\times$ baseline, 1.40$\times$ reference).

\paragraph{L2-025: Mixture-of-experts (MoE) expert-parallel backward.}
This task computes the backward pass for a 256-expert top-8 MoE with hidden dimension $H{=}4096$, intermediate dimension $I{=}2048$, FP32 data, and 16 input shapes with token count $T{=}2048$--$6144$. The reference loops over experts sequentially. Four parallel agents independently converged on counting-sort + batched GEMM in a single round. Using the benchmark's variable names, $W_g$, $W_u$, and $W_d$ are expert projection weights; $G$ is the incoming output gradient; and \texttt{dGate}, \texttt{dUp}, \texttt{dInter}, and \texttt{dX} are intermediate or input gradients. Key optimizations:

\begin{enumerate}
    \item \textbf{Padded-capacity layout + TensorFloat-32 (TF32).} Tokens are binned by expert via \texttt{atomicAdd} into $[256{\times}M_\text{pad}]$; all GEMMs run as cuBLAS TF32 strided-batched (4.5$\times$ faster than IEEE FP32, within tolerance).

    \item \textbf{Algebraic gradient reuse.} The product $G{\times}W_d$ is computed once and reused for both the router gradient and \texttt{dInter}, saving ${\sim}$8.6\,GB of HBM traffic.

    \item \textbf{Fused SiLU--top-$k$--SwiGLU backward.} One CUDA kernel computes the intermediate activation, routing gradient (warp reduction and scatter), and SwiGLU backward, eliminating three DRAM round-trips.

    \item \textbf{SM100 dual-GEMM scatter-atomic.} A single CUTLASS kernel with a $K$-concatenated mainloop (four TMA descriptors) fuses $\mathtt{dGate}\times W_g+\mathtt{dUp}\times W_u$ and scatters into \texttt{grad\_hidden} via FP32 vector atomics, eliminating the \texttt{dX} buffer. It reaches 7.05\,TB/s and replaces the slower cuBLAS-bmm alternative for this subproblem in the accepted campaign.

    \item \textbf{Host-mapped max-count fast path.} A fused histogram uses block-aggregated \texttt{atomicMax}; when mapped memory is available, the last block publishes via \texttt{\_\_thread\-fence\_system} to a host-visible slot, with a pinned-copy fallback otherwise.
\end{enumerate}
\noindent Final SOL: \textbf{0.535106} (1st\footnotemark[\value{leaderboardranknote}], 16/16 shapes faster than baseline, 1.13$\times$ baseline, 291$\times$ reference).

\begin{algorithm}[t]
\caption{L2-025 fused MoE backward}
\label{alg:l2-025-pseudocode}
\begin{algorithmic}[1]
\REQUIRE hidden states $X$, output gradient $G$, top-$k$ experts and weights, expert weights $W_g,W_u,W_d$
\STATE flatten token--expert assignments; count rows per expert with a fused histogram/max kernel
\STATE choose padded capacity $M_\text{pad}$ and place token--expert pairs into expert-contiguous rows
\STATE gather $X$ and $G$ into padded expert-major buffers, zero-filling gap rows
\STATE recompute gate and up activations with per-shape cuBLASLt TF32 batched GEMMs
\STATE compute $G_d \leftarrow G W_d$ once and reuse it for router and activation gradients
\STATE run one fused CUDA kernel for SiLU, top-$k$ gradient, and SwiGLU backward
\STATE compute gate/up/down weight gradients with batched GEMMs
\IF{$M_\text{pad} \leq 128$}
    \STATE run SM100 dual-GEMM CUTLASS kernel and scatter-atomic directly into \texttt{grad\_hidden}
\ELSE
    \STATE compute two input-gradient GEMMs and reduce/scatter with \texttt{sum\_hidden2}
\ENDIF
\RETURN gradients
\end{algorithmic}
\end{algorithm}

\paragraph{Quant-031: NVFP4 grouped-query attention (GQA).}
The task implements GQA (40 Q heads, 8 KV heads, $D{=}128$) using NVIDIA's NVFP4 4-bit floating-point format with E2M1 1${\times}$16 block scaling on Blackwell's fifth-generation Tensor Cores. The reference calls \texttt{torch.\_scaled\_mm} per batch.

\begin{enumerate}
    \item \textbf{Gen~1 (SOL~0.971): BF16 dequantization + graphs.} A fused Triton FP4-to-BF16 conversion feeds batched \texttt{torch.bmm} (FP4${\times}$E4M3 products are exact in BF16). CUDA Graph replay eliminates launch gaps.

    \item \textbf{Gen~2 (SOL~0.980): tcgen05 NVFP4.} Using CUTLASS~4.4.2's persistent block-scaled GEMM kernel via CuTe DSL, Triton emits packed FP4 nibble pairs and scale factors in a 512-byte atom layout; tcgen05 GEMMs use TMA loads and remain bit-identical to the BF16 path.

    \item \textbf{Gen~3 (SOL~0.988): Fused softmax--W quantization.} The causal mask forces softmax row\,0 $= [1,0,\ldots]$, so $\text{amax}(W){=}1.0$ \emph{exactly}. The scale chain and FP4 packing execute in registers inside the softmax kernel, eliminating the W-quantization DRAM pass (18--27\% of the span).
\end{enumerate}
\noindent A failed tcgen05-epilogue fusion (per-element scalar loops) caused 8--18$\times$ slowdowns---recorded as a trap. Final SOL: \textbf{0.988423} (1st\footnotemark[\value{leaderboardranknote}], 16/16 shapes faster than baseline, 43.8$\times$ baseline, 1327$\times$ reference).

\begin{algorithm}[t]
\caption{Quant-031 native NVFP4 grouped-query attention}
\label{alg:q031-pseudocode}
\begin{algorithmic}[1]
\REQUIRE query $Q$, key $K$, value $V$, causal mask, output buffers for attention output and weights
\STATE reshape heads into grouped-query batches
\STATE compute fused amax values for $Q,K,V$
\STATE quantize and pack $Q$ and $K$ into NVFP4 bytes with blocked scale factors
\STATE quantize/pack $V^\top$; for power-of-two sequence lengths, overlap this with $Q/K$ packing
\STATE run CuTe/CUTLASS SM100 block-scaled NVFP4 GEMM for raw scores $S \leftarrow QK^\top$
\STATE apply scale, causal mask, and softmax in Triton
\IF{sequence shape permits fused weight quantization}
    \STATE pack softmax weights to NVFP4 inside the softmax kernel
\ELSE
    \STATE quantize softmax weights in a separate packing kernel
\ENDIF
\STATE run second NVFP4 GEMM for $O \leftarrow \mathrm{softmax}(S)V$
\STATE optionally use CUDA graph replay for small launch-bound shapes
\RETURN
\end{algorithmic}
\end{algorithm}

\paragraph{FI-014: Paged prefill causal GQA.}
The task performs causal paged-attention prefill with 32 Q heads, 4 KV heads, $D{=}128$, page size 1, and 30 variable-length workloads. The reference is a na\"ive Python loop.

\begin{enumerate}
    \item \textbf{Gen~1 (SOL~0.946): CUDA multi-Q.} With \texttt{BLOCK\_Q}${=}2$, two query tokens share K/V loads, halving traffic; online softmax remains in registers, with warp-shuffle reductions and binary-search batch lookup.

    \item \textbf{Gen~2 (SOL~0.966): Triton flash-attn.} $[16{\times}128]$ tensor-core tiles; constexpr strides (shifts not multiplies), int32 indices, \texttt{evict\_first} KV policy. Batch-specialized fast paths eliminate compressed-sparse-row (CSR) lookups.

    \item \textbf{Gen~3 (SOL~0.986): split-causal cuDNN path for WL11.} For workload 11 (WL11; query length $S_q{=}4062$, key/value length $S_\text{kv}{=}16386$), the archived solution freshly gathers the first sequence's paged K/V into contiguous buffers, runs two \texttt{aten} cuDNN SDPA calls---one non-causal prefix and one causal tail---and merges the outputs and LSE values with a Triton rescaling kernel. No graph, plan, or input-derived buffer is cached across calls in this path; remaining tail sequences use the lean Triton paged-attention kernel.
\end{enumerate}
\noindent Final SOL: \textbf{0.986154} (1st\footnotemark[\value{leaderboardranknote}], 30/30 shapes faster than baseline, 143.8$\times$ baseline, 61{,}973$\times$ reference).

\begin{algorithm}[t]
\caption{FI-014 paged prefill causal GQA}
\label{alg:fi014-pseudocode}
\begin{algorithmic}[1]
\REQUIRE query rows, paged K/V cache, query and KV indptrs, page indices, scale, output and LSE buffers
\IF{there are no query rows}
    \STATE fill LSE with $-\infty$ and return
\ENDIF
\IF{KV cache is small and the workload is not tiny}
    \STATE run one lean Triton attention kernel over paged K/V rows
\ELSIF{KV cache is large}
    \STATE gather the long first sequence's paged K/V into contiguous buffers
    \STATE run cuDNN SDPA on non-causal prefix and causal tail
    \STATE merge the two outputs with LSE-rescaling in Triton
    \STATE run the lean Triton paged-attention kernel for the remaining tail sequences
\ELSE
    \STATE run the paged Triton BLOCK\_Q=2 kernel with shape-selected BLOCK\_N
\ENDIF
\RETURN
\end{algorithmic}
\end{algorithm}

\paragraph{L2-053: text decoder layer with self-attention and MLP.}
L2-053 is a text-decoder-layer task combining self-attention and an MLP. The solution uses Triton RMSNorm/RoPE/attention kernels, shape-gated fused or fallback INT8 gate/up projections, and a BF16 fallback path for smaller shapes or unsupported INT8 execution.

\begin{algorithm}[t]
\caption{L2-053 shape-gated decoder-layer fusion}
\label{alg:l2-053-pseudocode}
\begin{algorithmic}[1]
\REQUIRE hidden states, attention weights, MLP weights, RMSNorm weights, RoPE parameters, destination output
\STATE $M \leftarrow$ flattened batch--sequence token count
\IF{$M < 1500$}
    \STATE run fused RMSNorm plus K/V-weight packing; compute merged K/V projection
\ELSE
    \STATE run RMSNorm; compute Q, K, and V projections separately
\ENDIF
\STATE apply RoPE and split Q/K into attention layout
\IF{sequence length is 128}
    \STATE run custom single-tile Triton attention
\ELSE
    \STATE run PyTorch scaled-dot-product attention with GQA enabled
\ENDIF
\STATE compute output projection and add first residual
\IF{INT8 is supported and $M \geq 1000$}
    \STATE run fused add+RMSNorm+activation quantization
    \STATE quantize gate/up weights in one launch
    \IF{warp-specialized fused INT8 SwiGLU GEMM is available}
        \STATE compute gate/up GEMM with fused dequantization and SwiGLU epilogue
    \ELSE
        \STATE use \texttt{\_int\_mm} followed by a separate dequantization/SwiGLU kernel
    \ENDIF
\ELSE
    \STATE compute BF16 gate/up projections and a Triton SiLU-multiply kernel
\ENDIF
\STATE compute down projection and add the final residual into the destination
\RETURN
\end{algorithmic}
\end{algorithm}

\end{document}